**Title:** A High-frequency, Interaction-induced Pneumatic Oscillator Enabling Versatile Soft Robotics

**Authors:**
Longchuan Li[1], Shuqian He[2], Qiukai Qi[3]*, Ye Cui[4], Cong Yan[2], Kaige Jiang[5], Shuai Kang[4]*, Isao T. Tokuda[2], Zhongkui Wang[2], Shugen Ma[6], Huaping Liu[7]

**Affiliations:**
[1]College of Information Science and Technology, Beijing University of Chemical Technology, Beijing, China.
[2]Department of Mechanical Engineering, Ritsumeikan University, 1-1-1 Nojihigashi, Kusatsu, Shiga, Japan.
[3]Department of Engineering Mathematics, University of Bristol, Bristol.
[4]College of Mechanical and Electrical, Beijing University of Chemical Technology, Beijing, China.
[5]College of Materials Science and Engineering, Beijing University of Chemical Technology, Beijing, China.
[6]Thrust of Robotics and Autonomous Systems, The Hong Kong University of Science and Technology (Guangzhou), Guangzhou, China.
[7]Department of Computer Science and Technology, Beijing National Research Centre for Information Science and Technology, Tsinghua University, Beijing, China.
*Corresponding author. Email: qiukai.qi@bristol.ac.uk, kangshuai@buct.edu.cn

**Abstract:** Soft pneumatic robots are capable of diverse motion behaviors, yet their autonomy remains constrained by reliance on rigid valves and external control systems. Achieving intrinsic signal generation within soft materials is key to realizing fully soft, self-regulating machines. Here we present a High-frequency Interaction-induced Pneumatic Oscillator (HIPO) that harnesses internal mechanical–fluidic coupling to generate control signals without electronic or rigid components. The oscillator autonomously produces periodic pressure oscillations up to 97 Hz and directly drives actuation, enabling fast and continuous motion. We demonstrate its versatility through bio-inspired prototypes, including an insect-like crawler, a butterfly-like flapper, and a duck-like swimmer. This work introduces a design paradigm in which oscillation itself becomes motion, unifying signal generation and actuation within a single soft body, and opening pathways toward scalable, physical intelligence in soft robotic systems.

**Introduction**

Pneumatically actuated soft robots offer distinct advantages including low cost, high safety, environmental adaptability and biocompatibility[1, 2]. These systems realize diverse deformations—such as extension, twisting and bending—via fluidic actuation[3-6], enabling motion modes including crawling, jumping, swimming and grasping[7-11]. Such capabilities hold significant promise across medical, wearable and bioinspired applications[12-15]. However, the actuation of most existing pneumatic soft robots relies on rigid and bulky components such as solenoid valves and pressure regulators, resulting in systems with substantial rigid parts that hinder full soft integration[16]. This reliance not only increases structural complexity but also limits reliable operation in harsh environments with strong electromagnetic interference or high temperatures.

A central challenge in advancing fully soft robots lies in eliminating the dependence on external rigid control modules by leveraging the robot's intrinsic physical intelligence to generate control signals autonomously. In conventional robotics, signal generation is typically governed by rigid electronics, constraining flexibility and limiting adaptability in extreme settings. Micro pneumatic logic provides an effective approach for generating control signals, yet its capacity for regulating gas flow remains intrinsically limited[17]. In contrast, nature offers a wealth of physical and biological systems that achieve rhythmic behavior without electronics. Examples include vocal cords vibrating under airflow[18], leaves and flags fluttering via flow-induced instabilities[19, 20], Drosophila flapping at high-frequency using muscle elasticity and stretch activation[21], and the Venus flytrap snapping shut via structural bistability[22]. These phenomena highlight that efficient self-sustained oscillation often arises from finely tuned interactions among materials, structures, and fluid flows[23, 24]. Yet, achieving compact, high-frequency soft oscillators remains difficult due to the inherently low modulus, limited energy storage, and high internal dissipation of soft materials[25].

Existing soft oscillators often incorporate auxiliary elastic structures to overcome these limitations, such as slit hemispherical membranes[26-28] or bistable valves[29, 30]. These designs depend on mechanical instability in added structures rather than on intrinsic dynamic interactions among inherent soft components. An alternative strategy exploits interactions between soft pneumatic channels and airflow to induce autonomous flow switching and alternating actuation. Representative designs include frequency-tuned fluidic oscillators[31], tube-ball mechanisms[32], bistable textile structures[33], buckling plate ring oscillators[34], and programmable soft valves[35]. More recently, self-sustained oscillating thin tubes have been used for rapid autonomous motion in robots[36], though these too depend on the instability of specific tubing structures.

While these approaches have significantly advanced electronics-free pneumatic actuation, realizing oscillatory behavior within a fully soft body remains challenging due to two key limitations: (1) most pneumatic oscillators generate control signals but cannot directly drive motion; and (2) the high dissipation of soft materials restricts oscillation frequency and amplitude without rigid reinforcements. Consequently, a compact, high-frequency, and scalable soft oscillator based purely on intrinsic material interactions is still lacking.

Here, we introduce an interaction-induced oscillation strategy that leverages internal mechanical–fluidic coupling within soft materials to generate self-sustained oscillations. Based on this principle, we develop a compact High-frequency Interaction-induced Pneumatic Oscillator (HIPO) composed of a holed membrane and a flexible cover. Under constant airflow, periodic contact–separation between these two components establishes nonlinear boundary conditions that drive spontaneous oscillation. This mechanism enables: (1) oscillation-as-motion, where the cover directly interacts with the environment, achieving a fully soft oscillator with multimodal, scalable, and efficient motion; and (2) self-interactive oscillation, wherein simple internal interactions overcome material dissipation to convert steady airflow into oscillations up to 97 Hz (see Supplementary Movie 6).

We systematically analyze the internal dynamic interactions of HIPO, focusing on how interactions between the membrane and cover induce high-frequency oscillation. A simplified spring–damper model is developed to elucidate the energy transfer mechanisms during oscillation and reveal the nonlinear behaviors, especially the relationships among structural parameters, control inputs, and the resulting oscillation. The model clearly shows how the cover triggers oscillation and explains its nonlinear behavior, especially the relationship among structural parameters, control inputs, and the resulting oscillation. Finally, we demonstrate HIPO's versatility across diverse environments, achieving insect-like fast crawling (absolute speed of 50.27 cm/s), butterfly-like flapping flight, and duck-like water surface swimming, showcasing HIPO's potential in high-frequency, large-amplitude, and agile soft robots. We further construct an enhanced-frequency HIPO entirely from silicone elastomers, reaching 97 Hz oscillation, paving the way for applications in medical vibration therapy and microrobotics.

## Results

### Principles of HIPO

To achieve self-sustained oscillation, a system must integrate three essential elements: negative damping, positive feedback, and nonlinearity[37]. A familiar example is a child's swing: the external push provides negative damping, allowing the system to absorb energy from the environment; the positive feedback ensures that the applied force reinforces the periodic back-and-forth motion rather than inducing lateral or rotational deviations; and nonlinearity, such as the velocity-dependent increase in air resistance, limits the amplitude growth and stabilizes the oscillation.

Inspired by these principles, we observed that the inflation of a pneumatic chamber inherently embodies the first two elements—negative damping (energy input through airflow) and positive feedback (membrane expansion)—as shown in Fig. 1a, b. To introduce the required nonlinearity, we incorporated a small hole in the membrane and added a flexible cover capable of sealing it (see fabrication details in Supplementary Note 1 and Supplementary Fig. 1). The interaction between the holed membrane and the cover introduces a switch-like nonlinearity: when the cover contacts the membrane, it closes the hole; when it separates, the hole reopens.

The oscillation process described in this paper can be divided into three distinct stages, as shown in Fig. 1a: (i) Initial stage ($0 - a_{t1}$): The oscillator inflates, with the air inflow into the chamber exceeding the outflow. The internal pressure in the air chamber of the oscillator slowly increases, causing the membrane to expand; (ii) Contact stage ($a_{t1} - b_{t1}$): Once the membrane makes contact with the cover, the cover closes the hole, allowing only air inflow into the chamber. The internal pressure in the air chamber rapidly increases, causing the membrane to expand in a short period, which leads to the cover rotating counterclockwise. (iii) Separation stage ($b_{t1} - a_{t2}$): The cover rotates and separates from the membrane. During this stage, the expansion of the membrane enlarges the hole, allowing the air outflow to greatly exceed the inflow, which reduces the internal pressure $P_{covered}$ and causes the membrane to contract within a short period. Simultaneously, the cover rebounds clockwise to reclose the hole, due to the elasticity at the top connection, resetting the system's stage and entering the contact stage. This causes the membrane to expand again and pushes the cover to rotate counterclockwise. In Fig. 1b, we compare the inflation behaviors of the uncovered and covered actuators (Supplementary Movie 1), and the corresponding pressure variation is illustrated in Fig. 1c. The uncovered actuator remains invariant once the internal and external pressures reach equilibrium, whereas the covered actuator enables oscillation spontaneously. Once this pattern emerges, the system gradually settles into a periodic motion, in which the membrane and cover rapidly alternate between contact and separation, maintaining continuous self-sustained oscillation. During this process, the cover not only couples with the perforated membrane to generate feedback signals in Fig. 1d, but also transmits high-frequency motion to the external environment.

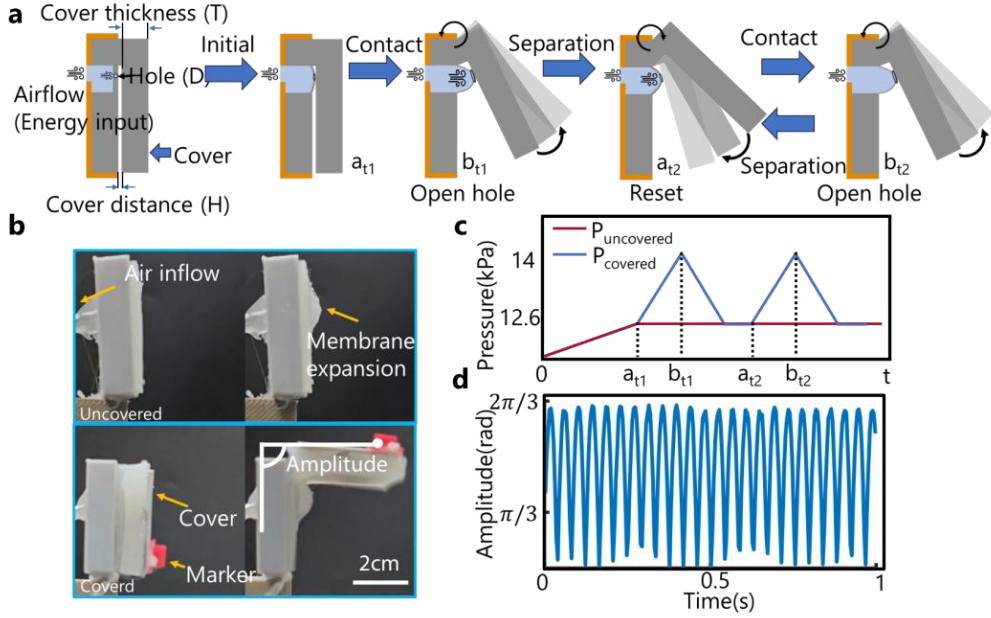

**Fig. 1. Principle of HIPO. (a)** Schematic diagram of the self-sustained oscillation principle. **(b)** Real images of the uncovered and covered oscillators. **(c)** Amplitude of the cover during the self-sustained oscillation process. **(d)** Schematic diagram of the internal chamber pressure variations in the air chamber of uncovered and covered actuators during self-sustained oscillation. (Pressure data measured see Supplementary Fig. 2).

**Simplified model of HIPO**

To verify whether the oscillation is induced by the interaction between the cover and the membrane, we developed a simplified dynamic model. To effectively capture the underlying dynamics and key factors influencing the oscillatory performance of HIPO, the membrane was abstracted as a ball with spring-damping characteristics, forming the ball-cover model (Fig. 2a). Following the classical spring–mass–damper representation of oscillators, the model is expressed as follows:

$$\begin{cases} M_{ball}\ddot{x}_{ball} = F_{in} + F_{out} - k_{ball}x_{ball} - c_{ball}\dot{x}_{ball} \\ I_{cover}\ddot{\theta}_{cover} = -k_{cover}\theta_{cover} - c_{cover}\dot{\theta}_{cover} \end{cases} \quad (1)$$

where $M_{ball}$ denotes the mass of the ball, $k_{ball}$ and $c_{ball}$ represent the translational spring stiffness and damping coefficient of the ball, respectively. Besides, $I_{cover}$ denotes the rotational inertia of the cover, $k_{cover}$ and $c_{overt}$ are the rotational spring stiffness and damping coefficient of the cover, respectively. $F_{in}$ is a constant force, denoting the air inflow of the oscillator. As the diameter of the hole increases with the expansion of the air chamber, the outflow of air increases, making the hole act as a source of aerodynamic resistance for the air chamber's expansion. Therefore, in the ball-cover model, as the ball displacement, $x_{ball}$ increases, a linear restoring force $F_{out}$ is assumed to generate aerodynamic resistance, similar to a spring's resistance to displacement.

$$F_{out} = \begin{cases} -S_D x_{ball} & x_{cover} - x_{ball} - x_{size} \geq 0 \\ 0 & x_{cover} - x_{ball} - x_{size} < 0 \end{cases} \qquad (2)$$

Here, $S_D$ represents the size of the hole. And $x_{size}$ is the distance between the centroid of the ball and the centroid of the cover, representing the deformation range of the soft material during the collision process. The collision is simplified as an elastic collision model (Supplementary Note 2). The $F_{out}$ introduces an interaction similar to the actual HIPO: when the cover collides with the ball, $F_{out}$ is turned off, and when the cover separates from the ball, $F_{out}$ is turned on.

Fig. 2b, c illustrates the self-sustained oscillation process of the ball-cover model, which can be divided into three phases similar to HIPO: (i) Initial phase ($0 - a_{t1}$): The ball accelerates under the influence of $F_{in}$ and makes contact with the cover at state $a_{t1}$, causing $F_{out}$ to close, and the ball transferring its kinetic energy to the cover through the collision; (ii) Contact phase ($a_{t1} - b_{t1}$): The ball continues transferring kinetic energy through the collision until state $b_{t1}$, when the ball separates from the cover due to its own elasticity, and $F_{out}$ reopens; (iii) Separation phase ($b_{t1} - a_{t2}$): Under the influence of $F_{out}$, the ball returns to its equilibrium position, while the cover rebounds clockwise due to its own elasticity and collides with the ball again at state $a_{t2}$. This collision resets the cover's phase, turning off $F_{out}$ and initiating the next cycle.

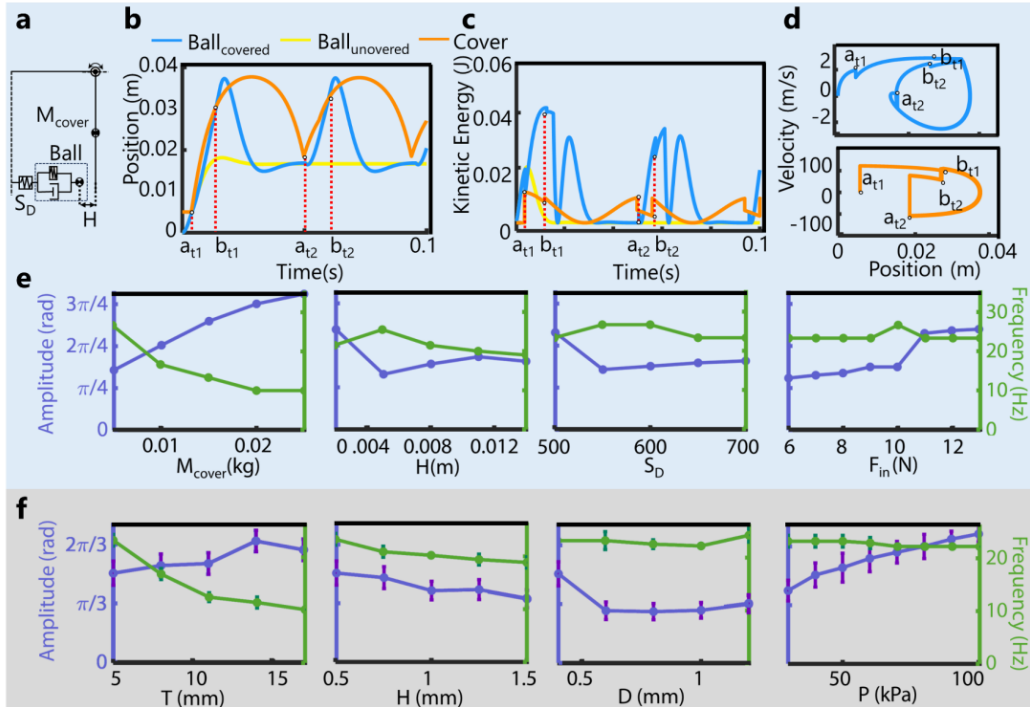

**Fig. 2. Simplified model of HIPO and experimental comparison. (a)** Schematic diagram of the simplified model of HIPO, reduced to the ball-cover model. **(b)** The displacement diagram of the ball$_{coverd}$, ball$_{uncoverd}$, and cover along the X-axis, with the cover's displacement calculated based on its rotation angle and the cover's length. **(c)** The kinetic energy changes of the ball and the cover during the oscillation process demonstrate how the collision transfers energy from the ball to the cover. This energy transfer compensates for the energy consumed by the cover due to damping and other dissipative forces. **(d)** Phase diagram of the ball$_{coverd}$ and cover motion. **(e)** The effect of different parameters on amplitude and frequency in the model. **(f)** The effect of different parameters on amplitude and frequency in the experiment.

As shown in the phase diagram in Fig. 2d, when the system enters the next oscillation cycle at $a_{t2}$, the motions of the cover and the ball asymptotically reach a limit-cycle. Consequently, the cover exhibits stable, periodic oscillations on the right side of its equilibrium position (see Supplementary Fig. 4), closely resembling the cover motion observed in the HIPO system in Fig. 1c. This further validates the consistency of the oscillatory behavior within the HIPO system. Fig. 2b and 2c show that the ball eventually reaches a constant output without the cover, demonstrating that the oscillation originates from the effective interaction between the holed membrane and the cover (Supplementary Movie 2). Since the cover of HIPO transmits high-frequency motion to the external environment, we regard the cover's oscillation as the primary output of the system. Thus, we focus on analyzing both the amplitude and frequency of the cover's oscillation.

To analyze the structural impact on HIPO performance, we defined the key geometric parameters shown in Fig. 1a, which are directly related to the structure that enables the interaction-induced self-sustained oscillation. Specifically, $T$ represents the cover thickness, which is used to uniformly reflect the cover mass $M_{cover}$; $D$ represents the diameter of the hole in the membrane, used to uniformly reflect the hole size $S_D$; and $H$ indicates the distance between the cover and the membrane. Additionally, we investigated the effect of the energy source on system performance through the inflow pressure $P$ and input force $F_{in}$. These variables allowed us to assess how both the physical structure and energy input influence the behavior and efficiency of HIPO system.

Fig. 2e illustrates the effect of different parameters on the amplitude and frequency of the cover in the model. An increase in cover mass $M_{cover}$ results in a decreasing in the oscillation frequency of the cover. This inverse relationship can be explained by considering the fundamental equation of harmonic oscillations, where the natural frequency $f$ is inversely proportional to the square root of mass $M$ ($f \propto \frac{1}{\sqrt{M}}$). As $M_{cover}$ increases, the system's moment of inertia increases, leading to a slower oscillation. Conversely, the amplitude of oscillation increases with $M_{cover}$, which can be attributed to the larger $M_{cover}$ absorbing more kinetic energy during collisions with the ball, thereby amplifying the oscillation.

Besides, the input force $F_{in}$ plays a significant role in controlling the amplitude of oscillation. With increasing input force, the amplitude rises substantially since the cover gains more energy. However, the influence of $F_{in}$ on the oscillation frequency is minimal. This is because the frequency of a self-sustained oscillation system is primarily determined by system's intrinsic properties, such as mass and stiffness, rather than external forces. This result is consistent with previous studies on harmonic oscillations, where the driving force significantly affects the amplitude, but the oscillation frequency remains largely constant, governed by the system's natural characteristics[38].

In contrast, both the cover distance $H$ and the hole size $S_D$ exhibit relatively minor effects on the amplitude and frequency of oscillation. The limited influence of $H$ can be attributed to the fact that this distance primarily affects the initial positioning of the cover rather than its dynamic properties once oscillation begins. Similarly, the hole size $S_D$ does not significantly alter the primary mechanical characteristics of the system. Small variations in $S_D$ may influence local pressure differences, but they do not substantially modify the global oscillation. These findings are consistent with fluid-structure interaction studies, where structural oscillation tends to be more sensitive to global mass and force changes rather than localized geometric variations[39].

Fig. 2f presents the effect of corresponding physical parameters of HIPO observed in actual experiments, with results closely aligning with the predictions from the model. This consistency suggests that our model accurately captures the key dynamics of HIPO, making it a reliable tool for understanding the mechanism and optimizing the system's behavior.

**HIPO-based insect-like fast-crawler**

Through a simple structural design, we developed a bio-inspired fast-crawler, drawing inspiration from the rapid locomotion of insects, particularly the escape behavior of cockroaches. The robot is powered by a HIPO integrated into its head, which is combined with a lightweight, 3D-printed body structure (Supplementary Fig. 5).

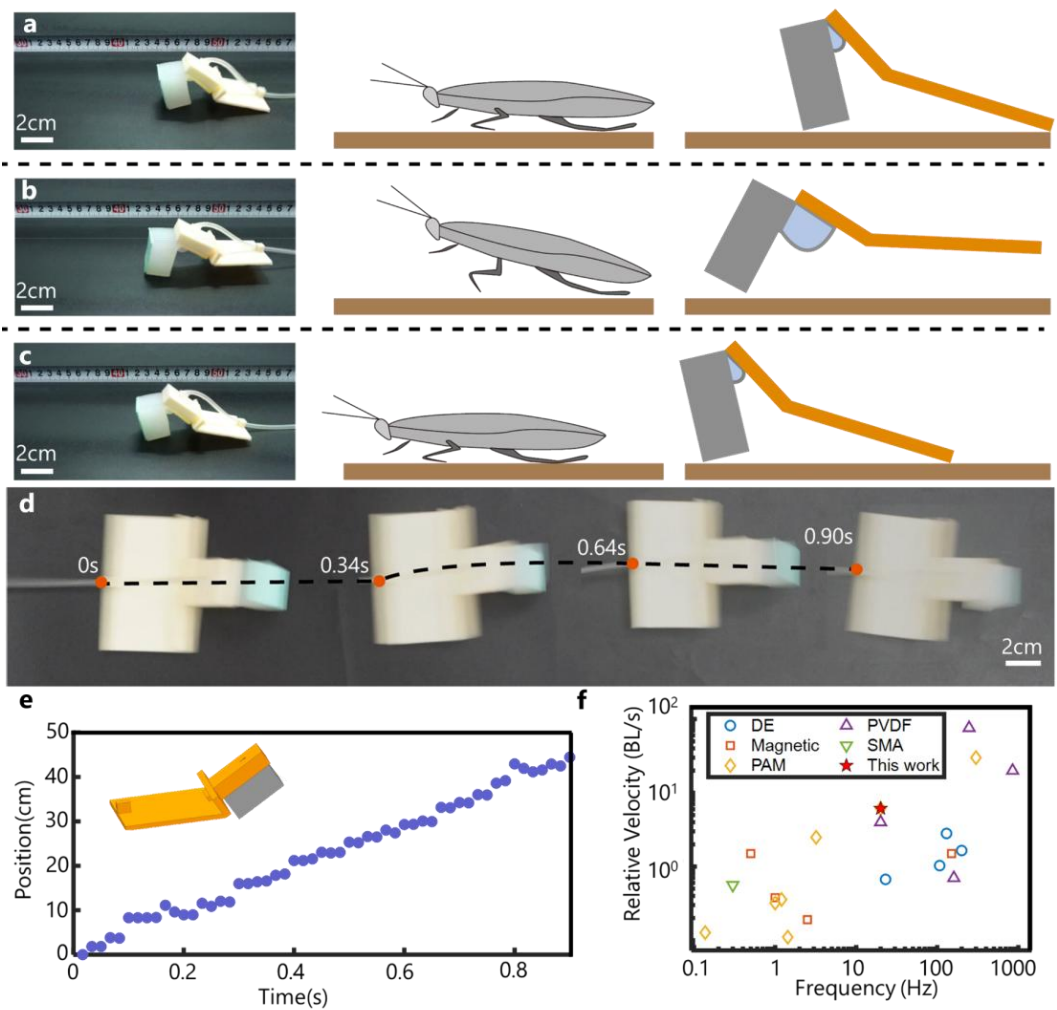

**Fig. 3. HIPO-based insect-like fast-crawler. (a)** The fast crawler prepares to jump, corresponding to the contact phase at state $a_{t1}$ in the limit-cycle of HIPO. **(b)** The membrane rapidly inflates, driving the body to unfold and propelling the robot into the air, corresponding to state $b_{t1}$ in the HIPO contact phase. **(c)** Upon landing, the robot's body folds toward the cover, pushing the entire robot forward and completing a crawling step, thereby preparing for the next phase of motion. **(d)** Top view of the fast-crawler's movement. **(e)** Time-displacement graph of a fast-crawler on the ground. **(f)** The relationship between the relative speed and body length of some soft mobile robots.

The robot's movement, as detailed in Fig. 3a-c (and shown in Supplementary Movie 3), is realized by oscillation generation and morphology design. Specifically, HIPO generates periodic forces, and the heavy cover creates large oscillation amplitude. This results in a high reactive force when the cover strikes the ground. This force produces a propulsive thrust that drives the robot's body to unfold. As a result, the robot transitions from a stable posture (Fig. 3a) to a jumping posture (Fig. 3b), mimicking the rapid escape jumps observed in cockroach species[40]. The robot then returns to the ground, where the rebound of the cover triggers the re-folding motion of the body. In addition, the HIPO-based silicone head has a higher friction coefficient and constitutes the majority of the mass, thus remaining nearly stationary during the folding stage. In contrast, the lighter 3D-printed body, with its lower friction coefficient, retracts toward the head under the elastic forces of the oscillator, resulting in forward crawling. This design takes advantage of the friction difference and mass distribution between the head and body, allowing the robot to achieve a crawling forward motion[41].

To characterize the driving performance of HIPO, we conducted a mobility experiment on a fast-crawler under a driving pressure of 90 kPa, as shown in Fig. 3d. Notably, the robot moved 45.2 cm in 0.9 seconds, achieving an impressive absolute speed of 50.27 cm/s, which corresponds to a relative speed of 6.13 BL/s (Fig. 3e). In Fig. 3f, we compared the movement speed of this robot with several representative soft crawlers[7, 33, 36, 42-56]. Despite its minimalistic structure, the HIPO-based crawler achieves rapid locomotion by combining high-frequency actuation with large stride amplitudes, outperforming conventional pneumatic robots in relative speed (Supplementary Table 1). While its absolute speed is slightly lower than that of some pneumatic oscillators, it still surpasses dielectric elastomer and piezoelectric systems, which are limited by low driving power and small strides. Overall, the HIPO-based robot demonstrates a significant advantage in absolute speed, fully leveraging the intrinsic characteristics of pneumatic actuators to achieve outstanding locomotion performance.

**HIPO-based butterfly-like flapper**

Traditional pneumatic robots are typically limited to locomotion on the ground or in water, with airborne motion remaining difficult to achieve. This limitation stems from constraints on actuation frequency and response speed, which hinder the generation of sufficient thrust for flight. In contrast, HIPO enables higher driving frequencies combined with larger deformation amplitudes, making it well-suited for flapping-wing applications.

To demonstrate the versatility of HIPO and its potential in flight dynamics, we designed a pneumatic bio-inspired flapper, drawing inspiration from the flight mechanics of butterflies (Supplementary Fig. 6). The structure of the flapping-wing robot is highly simplified, with two symmetrically arranged HIPOs forming the main body. The wings are constructed using a 3D-printed frame featuring a thicker leading edge and a thinner trailing edge, covered with PET film at the rear. These wings are connected to the covers of HIPO via thin linkages. As shown in Fig. 4a-c, the expansion of the air chambers of HIPO drives the covers, inducing a flapping. This motion introduces slight torsional deformation in the thin linkages, allowing the wings to execute a combined up-and-down flapping and twisting motion. This dual movement effectively replicates the wing-beating action of butterflies, generating aerodynamic thrust.

As shown in Fig. 4d, to further validate the driving performance of HIPO in the flapper, we conducted a rotational flight experiment at a driving pressure of 100 kPa. In this experiment, the robot was supported by a rod that allowed it to rotate around an axis (see Supplementary Movie 4). The flapping motion of the wings produced sufficient thrust to propel the robot in a circular flight path around the pivot. This setup provided a controlled environment to assess the effectiveness of HIPO as a driving mechanism for a lightweight flapping-wing robot, demonstrating its ability to generate the necessary forces for sustained flight behavior. The experimental results (Fig. 4e) showed that the robot completed 4 full rotations within 10 seconds. This experiment not only demonstrated the efficient driving capability of HIPO but also highlighted its potential applications in complex flight dynamics. It shows that the system can achieve relatively simple yet powerful flight performance, expanding the range of applications for pneumatic actuators.

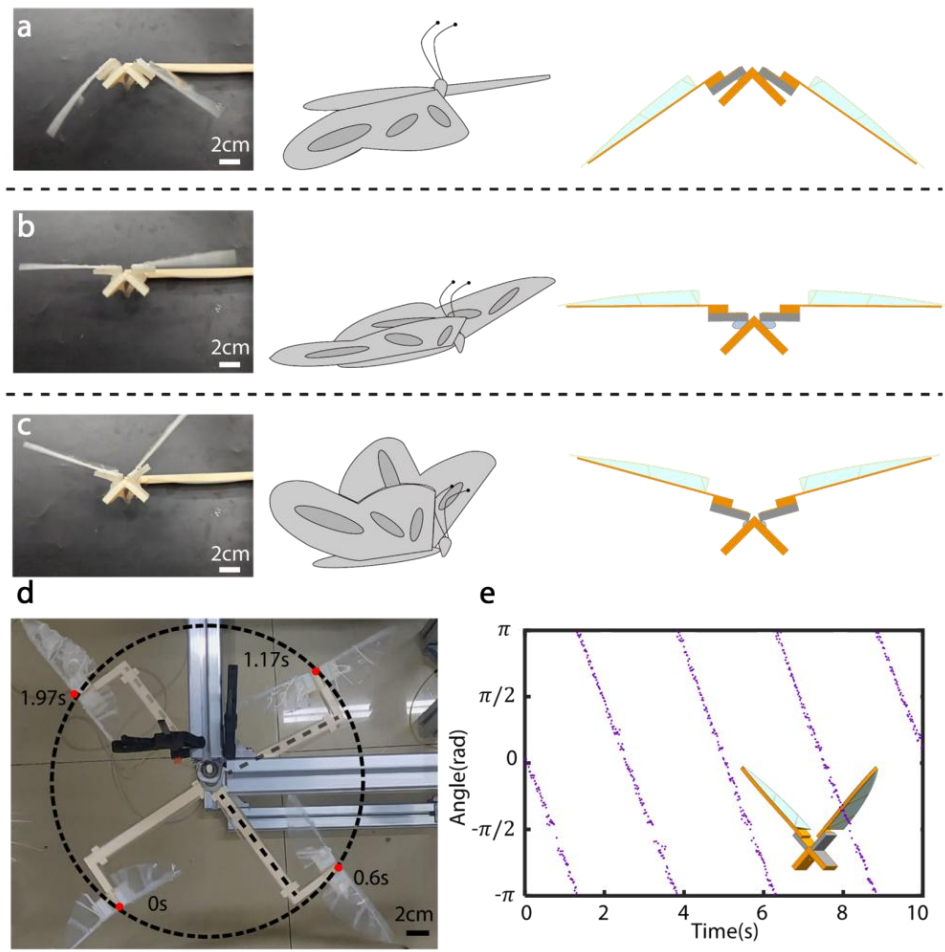

**Fig. 4. HIPO-based butterfly-like flapper. (a)** The wings are in the downstroke position, corresponding to the contact phase at stage $a_{t1}$ in the HIPO cycle. **(b)** The wings of the flapper are fully extended horizontally, corresponding to state $b_{t1}$ in the HIPO contact phase. **(c)** The wing reaches the upstroke position and subsequently performs a downward stroke, twisting to increase the angle of attack and generate thrust. **(d)** Top-down view of the butterfly-like flapper in rotational flight. **(e)** Time-angle graph of the butterfly-like flapper in rotational flight.

**HIPO-based duck-like swimmer**

While HIPO operates based on the principle of self-sustained oscillation and does not rely on external control signals, the system still retains maneuverability, which is a crucial performance factor in practical robotic applications. To demonstrate this capability, we drew inspiration from the propulsion mechanism of ducks, which push water backward to move forward, and designed a duck-like swimmer powered by HIPO.

The main body of the duck-like swimmer is designed in a boat shape, with HIPOs symmetrically arranged on both sides, inclined 45° downward (Supplementary Fig. 7). Fig. 5a-c shows that in this configuration, when HIPOs are activated, the covers rapidly strike the water surface, generating backward water flow that propels the robot forward. This setup efficiently converts oscillation into thrust, enabling smooth and continuous movement of the robot on the water surface.

To assess the maneuverability of the duck-like swimmer, we conducted experiments under different driving pressure conditions, with the results shown in Fig. 5d. Notably, we introduced a control experiment at 90 kPa where the system operated without covers. The results indicate that, although the airflow was directed backward due to the inclined arrangement of the HIPOs, this airflow alone is insufficient as an effective propulsion source. In contrast, when the covers were attached, HIPOs produced significantly increased thrust through high-frequency oscillation, demonstrates the conversion of airflow kinetic energy into the robot's motion, propelling the robot forward. At driving pressures of 40-70 kPa, the robot's movement speed increased significantly with the rising pressure. This is primarily because higher pressure amplifies the oscillation amplitude of HIPOs, allowing them to push the water more violently, generating stronger backward water flow and thus enhancing thrust. However, when the pressure reaches 80 kPa and 90 kPa, the speed increase plateaus, with little to no noticeable growth. This phenomenon can be attributed to the fact that as the amplitude continues to increase, the contact area and depth of the covers with the water increase during each downward stroke. This leads to greater water resistance, which offsets the additional thrust generated by the higher amplitude.

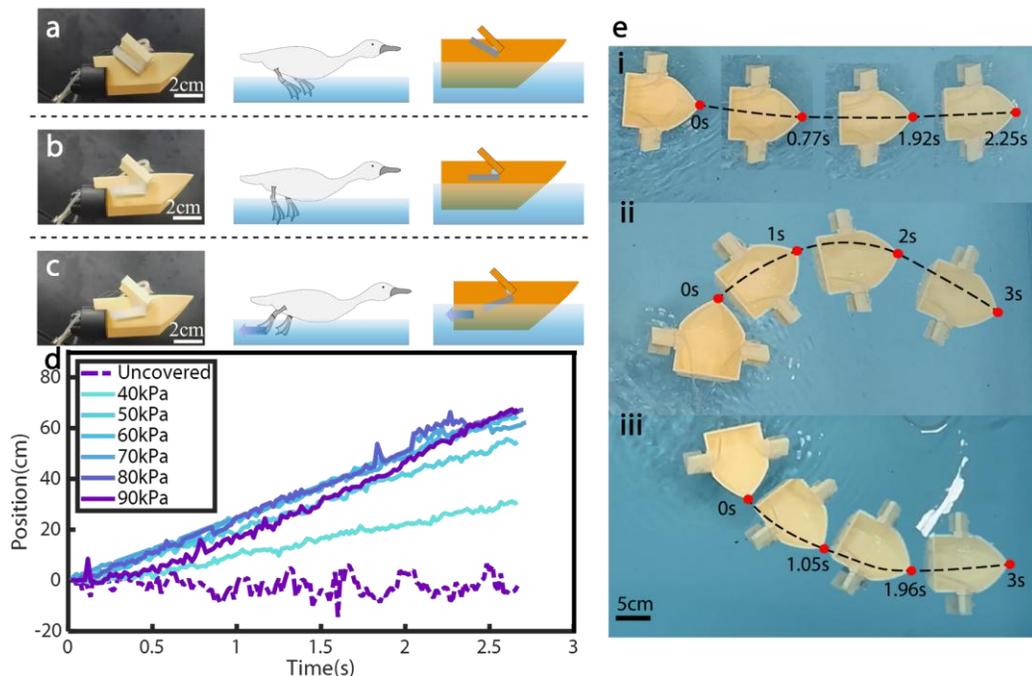

**Fig. 5. HIPO-based duck-like swimmer. (a)** HIPO is in the contact phase at state $a_{t1}$, with the cover not in contact with the water, and the system generates no thrust. **(b)** HIPO begins its downward stroke, with the cover making contact with the water, similar to how a duck's webbed feet push water backward. The interaction between the water and the cover

generates forward thrust. **(c)** HIPO completes its interaction with the water surface. At this point, the thrust temporarily decreases, the cover lifts out of the water, and the system enters a new actuation cycle, ensuring continuous propulsion. **(d)** Time-displacement graph of the duck-like swimmer under different driving pressures, including the motion of the robot without a cover at 90 kPa as a comparative. **(e)** Diagram of the duck-like swimmer's movement experiment on the water surface: **i**. Robot in straight-line motion; **ii**. Robot in right-turn motion; **iii**. Robot in left-turn motion.

In Fig. 5E, we present the experimental results of the duck-like swimmer performing straight-line movement, right turns, and left turns (Supplementary Movie 5) to assess its controllability. The results demonstrate that by adjusting the driving mode and air pressure input of HIPOs, the robot can control its direction and movement path to a certain extent. During straight-line movement, HIPOs on both sides operate at the same frequency and amplitude, generating balanced thrust that supports stable forward motion. For right and left turns, adjusting the air pressure on one side of HIPO creates an asymmetrical thrust distribution, enabling the robot to turn. Overall, the experiment preliminarily validates the robot's controllability in various movement modes, demonstrating potential flexibility and adaptability for real-world applications.

**Discussion**

Our research shows that by introducing internal structural interactions during the continuous inflation of the air chamber, a fully soft pneumatic system can successfully trigger high-frequency self-sustained oscillations. Compared with conventional pneumatic oscillators, the HIPO-based system requires no additional components and features a highly integrated and simplified structure, enabling extremely rapid response and superior motion performance.

Building upon this, we developed a ball-cover model that is in strong agreement with experimental observations, confirming that the oscillatory behavior originates from the interaction between the cover and the holed membrane, and explain the energy transfer mechanism of HIPO. This insight guided the design of HIPO module to achieve varying performance characteristics. Based on this, we implemented a high-performance fast-crawler and a flapper, as well as a maneuverable duck-like swimmer.

While HIPO has been demonstrated to achieve high-frequency oscillations and has been successfully applied in mobile robots across various scenarios, its potential as an oscillator extends well beyond these applications. Here, we have made structural adjustments and further developed an enhanced-frequency HIPO (Supplementary Fig. 8), increasing its frequency to approximately 97 Hz (see Supplementary Movie 6). The enhanced-frequency HIPO is entirely fabricated from biocompatible flexible silicone materials, offering a new perspective for innovation in medical devices. As shown in Fig. 6, the enhanced-frequency HIPO presents multiple advantages. First, its low oscillation intensity, which characterizes Low-Intensity Vibratory Stimulation (LIV). LIV can effectively promote bone mass increase and bone recovery, thereby contributing to improved skeletal health[57, 58]. Second, due to its compliance, the enhanced-frequency HIPO can safely interact with the human body, making it suitable for tissue regeneration therapy[59, 60]. Additionally, its high frequency characteristics enable it to generate high frequency airflow oscillations, therefore it can be applied in High-Frequency Chest Wall Oscillation (HFCWO) therapy to facilitate airway clearance and improve pulmonary function[61, 62]. On the other hand, the reduced size of the enhanced-frequency HIPO making it promising for application within the human body, such as in the intestinal environment, where compliant high frequency oscillations could stimulate intestinal peristalsis[63, 64].

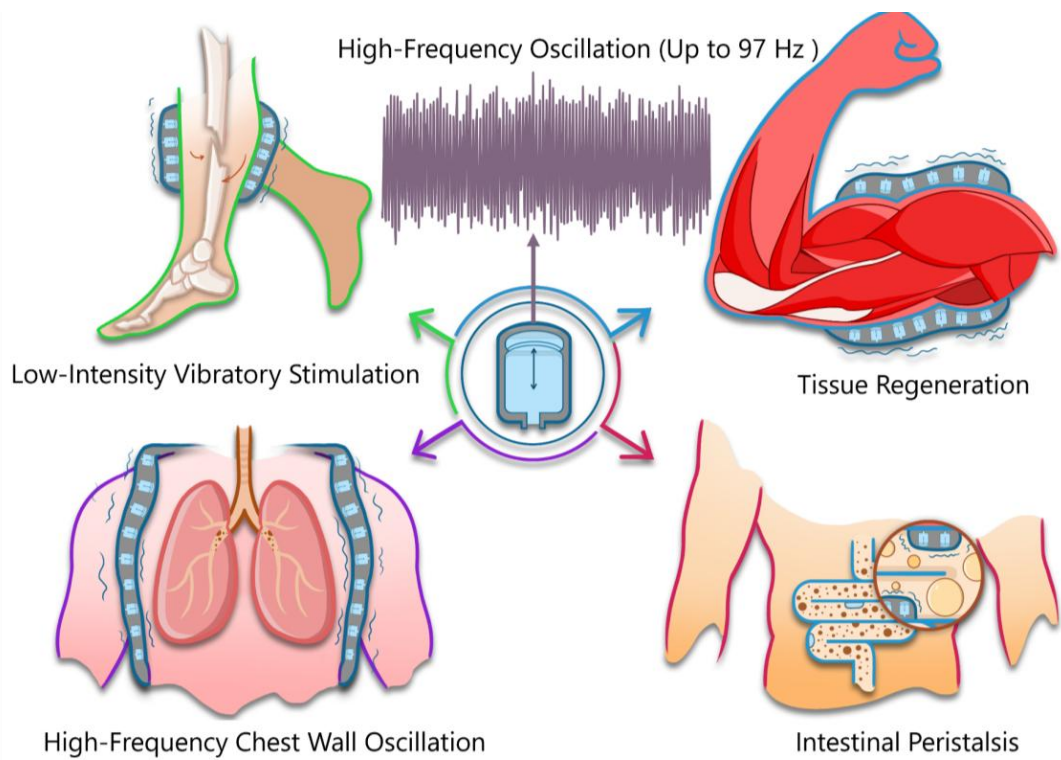

**Fig. 6. Enhanced-frequency HIPO with multiple application scenarios.** The enhanced-frequency HIPO, made entirely of soft silicone materials, exhibits good biocompatibility and holds great potential for application in various medical scenarios.

Although the current HIPO system operates at the centimeter scale and has yet to be experimentally validated at smaller dimensions, this limitation primarily arises from constraints in the laboratory setup. Since the core mechanism relies on the expansion dynamics of pneumatic chambers[65-67], miniaturization of the holed membranes and cover structures is theoretically achievable. This scalability suggests strong potential for micro-robotic applications, particularly in tasks demanding high-frequency oscillations and precise control. A miniaturized HIPO system could offer enhanced actuation flexibility and control resolution, further broadening its utility in micro-robotics.

**Methods**

**The fabrication process of HIPO**

The main body of HIPO consists of a silicone component featuring a cover and a holed membrane, integrated with 3D-printed parts for sealing the air chamber and connecting pneumatic tubing. The silicone part is fabricated using a casting method. The detailed fabrication process of HIPO is provided in Supplementary Note 1.

**Performance characterization of HIPO**

A high-frame-rate camera (ZV-E10, Sony Group Corporation, Japan) was used to record 10 seconds of footage at 120 frames per second. A red marker was added to the lower end of the cover, and the marker's position was processed in MATLAB. The middle 8 seconds of the recording were analyzed, with frequency calculated for each second, and the mean and

Page 12 of 17

variance were computed as experimental data. After increasing the cover's mass and lowering the oscillation frequency, the internal chamber pressure was measured using the MPX5100DP pressure sensor.

**Performance characterization of robots**

High-speed videos of different robots in motion were recorded at 120 frames per second using a high-frame-rate camera (ZV-E10, Sony Group Corporation, Japan). The footage was manually processed to isolate frames corresponding to different motion phases. The centroid position of the robot in each frame was then analyzed using MATLAB to obtain time-displacement data.

**Data availability**

All data necessary to evaluate the study are provided in the main text or Supplementary Materials. For any questions regarding experimental raw data, please contact Longchuan Li, Qiukai Qi, or Shuai Kang.

## Acknowledgments

**Funding:** This work was supported by Fundamental Research Funds for the Central Universities, China (buctrc202215), and the National Natural Science Foundation of China (62273340).

**Author contributions**
Conceptualization: L.L., S.H., Q.Q., S.K.
Methodology: L.L., S.H., Q.Q.
Investigation: L.L., S.H., Q.Q., Y.C., C.Y., K.J.
Visualization: S.H., Y.C., K.J.
Funding acquisition: L.L., S.K.
Project administration: L.L., Q.Q., C.Y., S.K., I.T.T., Z.W., S.M., H.L.
Supervision: Q.Q., S.K., I.T.T., Z.W., S.M., H.L.
Writing – original draft: L.L., S.H.
Writing – review & editing: L.L., S.H., Q.Q., Y.C., C.Y., K.J., S.K., I.T.T., Z.W., S.M., H.L.

## Ethics declarations
Competing interests: Authors declare that they have no competing interests.

**Supplementary Materials**